# Formality Style Transfer in Persian

Parastoo Falakaflaki[1], Mehrnoush Shamsfard[2*]

[1,2]Faculty of Computer Science and Engineering, Shahid Beheshti University, Tehran, Iran
[1] School of Electronic Engineering and Computer Science, Queen Mary University of London, UK
*Corresponding Author

## Abstract

This study explores the formality style transfer in Persian, particularly relevant in the face of the increasing prevalence of informal language on digital platforms, which poses challenges for existing Natural Language Processing (NLP) tools. The aim is to transform informal text into formal while retaining the original meaning, addressing both lexical and syntactic differences. We introduce a novel model, Fa-BERT2BERT, based on the Fa-BERT architecture, incorporating consistency learning and gradient-based dynamic weighting. This approach improves the model's understanding of syntactic variations, balancing loss components effectively during training. Our evaluation of Fa-BERT2BERT against existing methods employs new metrics designed to accurately measure syntactic and stylistic changes. Results demonstrate our model's superior performance over traditional techniques across various metrics, including BLEU, BERT score, Rouge-l, and proposed metrics underscoring its ability to adeptly navigate the complexities of Persian language style transfer. This study significantly contributes to Persian language processing by enhancing the accuracy and functionality of NLP models and thereby supports the development of more efficient and reliable NLP applications, capable of handling language style transformation effectively, thereby streamlining content moderation, enhancing data mining results, and facilitating cross-cultural communication.

## 1. Introduction

Advancements in neural networks for sentence generation now enable the transformation of an input sentence's formality style without compromising its inherent meaning, a concept known as formality style transfer [1]. This task involves altering the formality of a sentence while preserving its original meaning. The Persian language presents a particularly interesting case study for this task, given the substantial differences between its formal and informal styles.

Formal Persian serves as the linguistic standard in various institutional contexts, including academia, journalism, and scientific communication. It adheres to contemporary Persian grammar rules and is also used to instruct non-native speakers. On the other hand, informal Persian thrives in casual settings such as social media, everyday conversations, and storytelling. This mode of expression is characterized by unique elements like idiomatic phrases, colloquial expressions, and more relaxed grammatical structures and is often referred to as "shekaste-nevisi," which can be translated as "broken-writing", highlighting its simplified and shortened version of formal terms [2]. The informal Persian language has lexical, phonological, morphological and syntactic differences with its formal correspondence.

The increasing prevalence of informal language in digital platforms presents a significant challenge for existing NLP tools, which are generally optimized for formal text. This disconnect can lead to inaccuracies in text analysis, sentiment analysis, and even machine translation. As digital communication continues to rise, particularly in settings that traditionally required formal language such as customer service or corporate announcements, there's a growing need to bridge this gap. Developing models capable of converting informal text to its formal equivalent can not only improve the performance of various NLP applications but also expand their utility across a broader range of contexts. The ability to automatically formalize text could streamline content moderation, enhance data mining results, and even facilitate more effective cross-cultural communication.

However, the task is fraught with challenges, especially when it comes to the Persian language. Compared to well-resourced languages like English, Persian has fewer pretrained models and a limited corpus of annotated data. Traditional evaluation metrics like BLEU and ROUGE-L often fall short, potentially providing misleading assessments. For instance, these metrics can erroneously rate a rule-based model as highly effective, even when it merely duplicates the input text. This issue arises from the high frequency of words with identical formal and informal forms. Moreover, the abundance of such words increases the risk of model overfitting, causing models to merely replicate the input without making meaningful transformations. The task is further complicated by the necessity for syntactic adjustments and context-aware transformations, since the appropriate formal expression of a word can differ based on its surrounding context.

Our study aims to address these challenges comprehensively, focusing not just on lexical but also on syntactic modifications. Built on the Fa-BERT [3] architecture, we propose Fa-BERT2BERT to execute formality style transfer in Persian. A pioneering feature of our research is the incorporation of consistency learning, which leverages perturbed input sequences to enhance the model's comprehension of syntactic variations. We further augment this with gradient-based dynamic weighting to optimally balance loss components, thus allowing the model to better adapt to the complexities inherent in formalizing Persian text. We compare the efficacy of Fa-BERT2BERT against existing methodologies and introduce a new evaluative metric specifically designed to measure lexical and stylistic transformation.

In forthcoming sections, we will delve into existing literature on style transfer, surveying both general and formality-specific style transfer approaches across multiple languages, with a special emphasis on Persian. The methodology section will outline a detailed account of our innovative approach. The evaluation section provides details on our baseline models and evaluative metrics. Finally, the results section will present our evaluation results, compare Fa-BERT2BERT's performance against existing methods, and offer an in-depth analysis of our findings.

## 2. Related work

Text style transfer (TST) aims to generate grammatically correct texts that convey desired content in a particular style. The conventional seq2seq architecture [4] uses encoders to compress input into a vector and decoders to reconstruct text, but this method is limited by information loss and inability to consider the varying significance of input segments. The introduction of attention mechanism [5] has mitigated these issues by allowing the model to focus on relevant parts of the input during the generation process.

Furthermore, the traditional seq2seq architecture requires extensive parallel data for training, posing a challenge in face of limited data. To address this, researchers have explored various innovative techniques to utilize the data more effectively. The Delete-Retrieve-Generate framework takes a unique approach by separating and substituting text segments based on style, which is then synthesized into coherent output by a seq2seq model [6]. Back-translation method disentangles style and content implicitly through converting text to an intermediary language and back, incorporating the target style during retranslation [7].

Xu et al. [8] employed a hybrid model that combines seq2seq for English style transformation with a style classification component, enabling bidirectional formality transfer and enhanced data efficiency. This model uses various losses during training, including a classifier-guided loss from the classification component, which is less costly due to its reliance on simpler formality annotations.

Lastly, Lieu et al. [9] introduced a semi-supervised framework that maximizes the utility of unlabeled sentences on the source side through consistency training. This involves generating pseudo-parallel data by perturbing source sentences with techniques like word deletion, synonym replacement, and spelling errors, then training the model to produce consistent results across these variations.

Despite notable progress in formality style transfer in high resourced languages like English, research in the field of Persian remains underdeveloped due to linguistic complexities, a lack of extensive annotated datasets, and the absence of powerful pretrained large language models (LLMs). Consequently, Persian is classified as a low-resourced language, characterized by limited resources in terms of datasets and pretrained language models.

A frequently used method, particularly in the context of the Persian, revolves around devising conversion rules tailored for transferring formality in text through a word-by-word conversion process. Notably, Armin and Shamsfard [10] have introduced an innovative approach based on N-grams, effectively integrating rule-based and statistical models tailored to the Persian. Naemi et al [11] categorized Persian informal words into multiple categories based on the amount of changes needed to make them formal. Then they find a list of formal candidates using statistical analyses combined with spell correction. Additionally, the Hazm library [12] offers a rule-based model for Persian, which we used as one of our baselines.

While the widespread used rule-based methods may be capable of making lexical adjustments to achieve formality style transfer, they are fraught with limitations. Firstly, it is impractical to devise rules that account for all the different variations and nuances involved in converting text from an informal to a formal style, considering the vastness and ever-evolving nature of language. Secondly, these methods often lack context-awareness, making them unsuitable for sentences where the meaning—and therefore the formal

equivalent— is context-dependent. Thirdly, these techniques usually fall short when there is a need to restructure the word order in the text according to contemporary Persian grammar.

Kabiri et al. [13] work with the Informal Persian Universal Dependency Treebank (iPerUDT) underscores the gap between formal and informal language processing, with dependency parsers showing reduced accuracy on informal data due to its unique linguistic features. This highlights the broader challenges faced in adapting computational models to informal Persian.

Rasooli et al. [14] addressed some of these challenges using a seq2seq transformer model with a BERT-based six-layer encoder-decoder that discerns between standard and colloquial Persian using language ID embeddings. Their model was pretrained on datasets from Wikipedia in Persian, Arabic, and English, utilizing a SentencePiece vocabulary consisting of 60,000 tokens.

Our research focuses on creating an innovative model for Persian formality style transfer, designed to adeptly manage both lexical and grammatical modifications. Furthermore, we introduce a new metric specifically crafted to assess the model's proficiency in converting words with different formal and informal forms and addressing the necessary grammatical adjustments.

## 3. ParsMap: The Informal - Formal Dataset

Our study employs ParsMap, an Informal-Formal Persian Corpus [2], including 50,000 annotated sentence pairs that cover both informal and formal styles of Persian. Figure 1 displays the diverse range of data sources utilized in the ParsMap corpus, which includes sentences from academic papers, social media comments, and customer reviews found on e-commerce websites. Some sentences were also crafted by linguistic experts to ensure quality and diversity. Each entry in the corpus contains an informal sentence, its formal equivalent created by a linguistic expert, the source of the informal sentence and the detailed word-level and phrase-level alignments between the two.

As discussed before, one of the primary challenges in this task lies in converting informal words with different formal forms. Approximately 45.8% of the words in the dataset's informal vocabulary have non-equivalent formal forms. Nearly 49.7% of the data entries exhibit divergent grammatical structures between the informal and formal sentences. This variation can be attributed to the more flexible structure of informal sentences as opposed to the conventional SOV (Subject Object Verb) format adhered to in formal Persian grammar. Table 1 provides a more comprehensive overview of the dataset's linguistic characteristics.

Upon closer inspection of word alignments, the ParsMap dataset features 49,560 unique informal-formal word pairings. The most frequently occurring pairs include رو (pr: ro) aligned with را (pr: rɑː) (the object marker), یه (pr: je) aligned with یک (pr: jek; tr: one[1]), and تو (pr: too) aligned with در (pr: dar; tr: in). These pairs appear in 7,367, 5,797, and 4,419 entries respectively. The dataset also uncovers a layer of linguistic complexity, revealing that 3,000 informal words have multiple formal equivalents. While some of these equivalents are synonyms, the choice of formal word can also be context-dependent. For example, the word

---

[1] To make the Persian examples understandable for non-Persian speakers, the pronunciation (pr:) and the English translation (tr:) will follow each Persian example throughout the paper.

تو can have varied meanings and pronunciations—either "you" (to) or "in" (too)—each correlating with a different formal word.

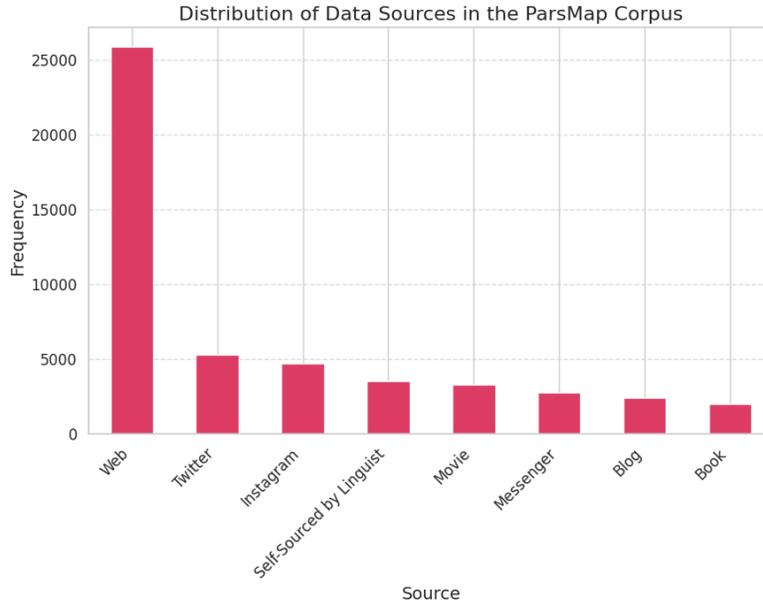

**Fig 1** Distribution of Data Sources in ParsMap Corpus

For training and evaluation, we allocated 80% of the dataset for training and the remaining 20% for testing. Figure 2 shows the distribution of input length across the entire test set. To evaluate effects of input length on model performances, in addition to evaluating the performance on the aggregated test set, we split the test data to two sets of short and long sets containing inputs with less than or greater than 11 tokens respectively.

**Table 1** Summary Statistics Comparing Linguistic Features of Informal and Formal Sentences in ParsMap

| Corpus Analysis Metric | Value |
|---|---|
| Average token count in informal sentences | 16.28 |
| Average token count in formal sentences | 16.18 |
| Percentage of words with distinct informal-formal forms | 45.81% |
| Percentage of sentences with different grammatical structures | 49.71% |

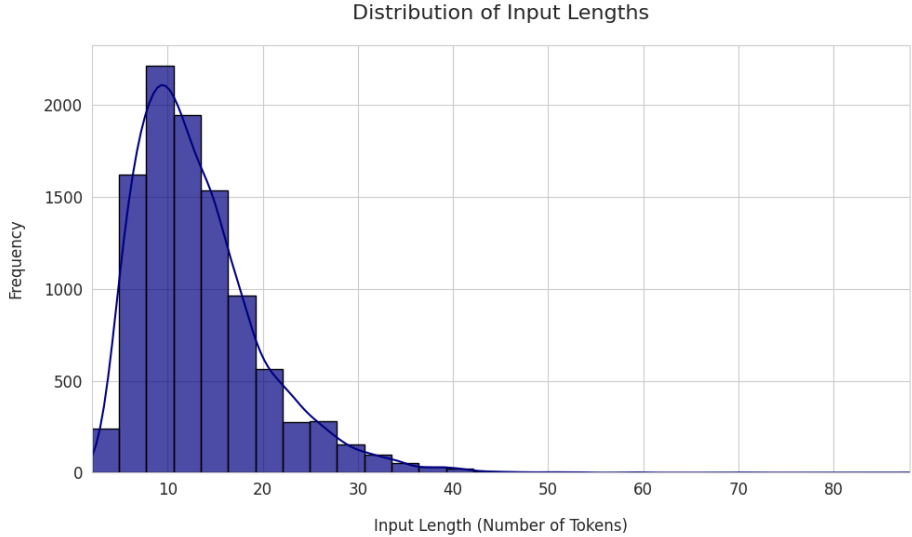

**Fig 2** Distribution informal inputs' length in terms of number of tokens in entire dataset

## 4. Proposed Model

In this study, we propose the novel Fa-BERT2BERT model built upon a transformer-based Seq2Seq architecture. As previously highlighted by Rothe et al. [15], initializing Seq2Seq models with pre-trained encoder checkpoints like BERT [16] has shown to yield robust performance without requiring exhaustive pretraining.

We employed the Fa-BERT [3] model as both the encoder and the decoder (FaBERT2BERT), which we subsequently fine-tuned on our dataset. FaBERT is a Persian BERT-base model pre-trained on the HmBlogs corpus [17], encompassing both informal and formal Persian texts. FaBERT, with 124 million parameters, replicates the original architecture of BERT with 12 hidden layers, each comprising 12 self-attention heads.

For training we experiment with two different approaches. The first method as outlined in 4.1.1 uses a dictionary for simple word replacement and then uses the transformer-based model for grammatical changes, while the second method (4.1.2) uses the transformer-based model as a context-sensitive, resource-independent style convertor, for both lexical and grammatical transitions.

### 4.1 Word Alignment Dictionary Strategy

In the first setting, we integrate a word alignment dictionary to replace informal words with their formal equivalents. Given that the dictionary contains multiple equivalents for some terms, the most frequent word is selected. This serves as a context-blind lexical style conversion, which we acknowledge as a limitation. Post replacement, the transformer-based model is trained following the training strategy outlined in 4.1.3.

## 4.2 Autonomous Approach

In this setting, we fully harness the capabilities of the pretrained transformer to manage both lexical and syntactic style conversions. Unlike the first setting, which employs a word alignment dictionary for lexical substitutions, this approach does not utilize any external resources. The model leverages its pretrained semantics and syntactic understanding to conduct context-aware conversions. By doing so, it can dynamically adapt to the complexities and nuances of different sentences.

## 4.3 Training Strategy

Considering the inherent flexibility of word order in informal Persian, it's imperative to address grammatical transitions when performing formality style transfer, to align with contemporary formal grammar. In response, we extend our model with a consistency learning approach. In addition to the standard loss functions, an auxiliary consistency loss is introduced as a regularization term. This loss is computed through the Mean Squared Error (MSE) between the outputs for input and its perturbed version, where two randomly selected words are swapped. For dynamic optimization, our model incorporates Gradient-based Dynamic Weighting which adjusts the weight for each loss based on its gradient during training, facilitating a more effective learning and adaptation to task complexity. We evaluated the effectiveness of this strategy by comparing the performance of the model trained using this approach against the same model architecture trained under standard setting.

Figure 3 illustrates our training strategy. In the supervised learning section, informal text input x1 is transformed by the transformer-based model (which can be either Pars-BERT2BERT or Fa-BERT2BERT) into formal output y'1, with the supervised loss calculated between the generated output y'1 and the true formal output y1. The consistency learning section demonstrates how perturbation methods are applied to the input before being processed by the same model to produce a perturbed output y''1. The Mean Squared Error (MSE) loss is then computed between the original and perturbed outputs to encourage consistency in the model's predictions. The final loss is a weighted sum of the supervised and consistency losses, with the weights α1 and α2 dynamically adjusted during training to optimize the model's performance.

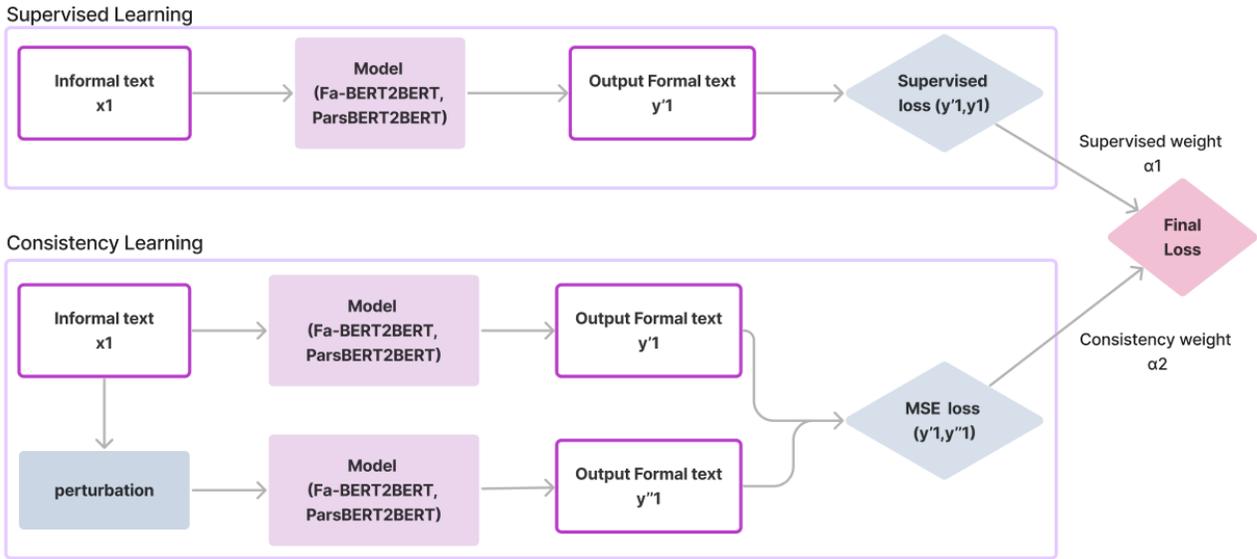

**Fig 3** The proposed training process, which incorporates both supervised and consistency learning methods enhanced by gradient-based dynamic weighting.

# 5. Evaluation

## 5.1 Baseline Models

To establish a performance benchmark for our proposed method, we employed several baseline models, each representing a distinct approach to the formality style transfer in Persian text. Below we explain these baseline techniques:

### 5.1.1. Rule-Based Models using Hazm Library

The Hazm library [12] is a Python toolkit designed for processing Persian text. For this study, we utilized Hazm's built-in formality converter, which is capable of converting informal text to its formal counterpart using rule-based methods.

### 5.1.2 Extended Rule-Based Model of Armin and Shamsfard

We augmented the rule-based model proposed by Armin and Shamsfard [10] by incorporating additional rules designed to better handle the complexities of Persian grammar. The workflow of the enhanced model is as follows:

Initially, sentences are tokenized into individual words based on whitespace. Predefined rules are then applied to identify a formal equivalent for each word. These rules generate a list of possible formal equivalents for each word, varying from none to multiple options. Table 2 shows some of these rules, providing examples for respective conversions. The subsequent conversion process depends on the number of potential equivalents identified, as described below:

- Single Equivalent: If a unique formal equivalent exists for a word based on the predefined rules, it is directly replaced. For instance, "می تونم" (pr: mi-tunam, tr: I can) is converted to "می توانم" (pr: mi-tava:nam).

- Multiple Equivalents: When multiple formal equivalents are possible for a word, the 'Best Equivalent Search Method' is employed to ascertain the most likely match. For example, "خونم" (pr: khūnam) could be either "خانه‌ام" (pr: kha:ne-am, tr: my home) or "خون من" (pr: khūn-e-man, tr: my blood). To resolve this ambiguity, a trigram consisting of the word and its two preceding words is formed and searched in a large corpus. The most frequently occurring trigram is selected, providing a context-sensitive, semantically appropriate translation.

- No Equivalent: In cases where no formal equivalent can be identified, the word is randomly splitted, and the model re-attempts to find formal equivalents for each fragment. For example, the word "نمیییییدونم" (Pronunciation: nemiiiii-dūnam, Translation: I don't know) is split into "نمیییی" (Pronunciation: nemiiiii) and "دونم" (Pronunciation: dūnam), which are then individually converted to "نمی" (nemi) and "دانم" (dānam), respectively.

**Table 2** Examples of conversion rules defined in 4.1.2

| Rule | Example | |
|---|---|---|
| | Informal | Formal |
| Adjective or Noun + plural marker "ا" (pr: 'a:') → Adjective or Noun + plural marker "ها" (pr: 'ha') | باغا (pr: 'ba:gha:', tr: 'gardens') | باغ ها (pr: 'ba:gh ha:', tr: 'gardens') |
| Adjective or Noun + informal clitic for the 3P plural copula "ن" (pr: 'n') → Adjective or Noun + formal 3P plural copula "هستند" (pr: 'hastand') | قشنگن (pr: 'ghashangan', tr: 'they are beautiful') | قشنگ هستند (pr: 'ghashang hastand', tr: 'they are beautiful') |
| Pronoun or Noun + 3P informal possessive pronoun clitic "شون" (pr: 'shoon') → Pronoun or Noun + 3P formal possessive pronoun clitic "شان" (pr: 'shān') | انرژی شون (pr: 'enerzhi shun', tr: 'their energy') | انرژیشان (pr: 'enerzhi sha:n', tr: 'their energy') |
| Direct Object + informal direct object marker "رو" (pr: 'ro') → Direct Object + formal direct object marker "را" (pr: 'rā') | کتابش رو (pr: 'keta:besh ro', tr: 'their book') | کتابش را (pr: 'keta:besh ra:', tr: 'their book') |
| Pronoun or Noun + informal direct object marker enclitic "و" (pr: 'o') → Pronoun or Noun + formal direct object marker enclitic "را" (pr: 'rā') | کتاب منو (pr: 'ketabe mano', tr: 'my book') | کتاب من را (pr: 'ketabe man ra:', tr: 'my book') |

### 5.1.3 Back Translation based model

Back-translation in neural machine translation (NMT) has been a widely adopted technique to produce synthetic parallel corpora, especially in the context of text style transfer [7]. In our experiment, as illustrated in figure 4, we utilized large mT5-based [18] models to perform bi-directional translations between English and Persian. We employed mt5-large-parsinlu-translation-fa-en[2] model to informal Persian texts into English. These were then back-translated to Persian using the mt5-large-parsinlu-translation-en-fa[3] model, which specifically generates formal Persian translation, aiming to generate outputs that are both content-consistent and formal.

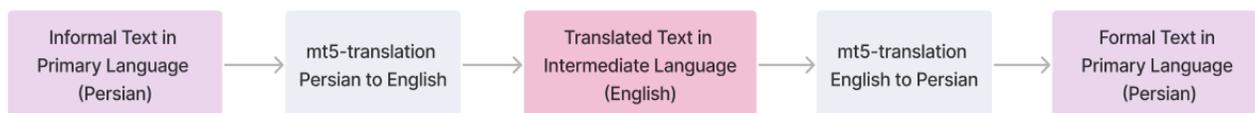

**Fig 4** A two-step back-translation process using mT5-based models for converting informal Persian text into a formal style. Initially, the text is translated from Persian to English, and subsequently, back-translated from English to Persian, with the end goal of preserving the original content while transferring the text style to formal Persian.

### 5.1.4 ParsBERT

In our primary model architecture, we initially utilized Fa-BERT as both the encoder and decoder to establish a baseline for performance. To further explore and assess the effectiveness of different models in our setup, we subsequently incorporated ParsBERT [19] in place of Fa-BERT, maintaining the same settings and model configuration (Pars-BERT2BERT with word alignment dictionary and Autonomous Pars-BERT2BERT). Specifically, we employ the third iteration of ParsBERT, a pre-trained monolingual model tailored for the Persian language corpus. ParsBERT adheres to the BERT-base specifications, which includes 12 hidden layers, 12 attention heads, and 768 hidden dimensions, amounting to a total of 110 million parameters.

## 5.2 Evaluation Metrics

Common metrics like BLEU (Bilingual Evaluation Understudy) [20] and ROUGE (Recall-Oriented Understudy for Gisting Evaluation) [21] are widely adopted for natural language generation (NLG) tasks. BLEU, typically utilized in machine translation evaluation, emphasizes precision, calculating the geometric mean of the precision of n-grams in the generated text against one or multiple reference texts. On the other hand, multiple variants of ROUGE, including ROUGE-N, ROUGE-L, and ROGUE-S, consider different textual overlaps like n-gram matches, longest common subsequence, and skip-bigram, respectively.

---

[2] [persiannlp/mt5-large-parsinlu-opus-translation_fa_en · Hugging Face](persiannlp/mt5-large-parsinlu-opus-translation_fa_en · Hugging Face)

[3] [persiannlp/mt5-large-parsinlu-translation_en_fa · Hugging Face](persiannlp/mt5-large-parsinlu-translation_en_fa · Hugging Face)

In the context of Persian formality style transfer, however, these metrics can be misleading as many words retain identical form in both formal and informal styles. For example, a model might achieve a high BLEU score by merely replicating input words without making any meaningful stylistic conversions. Likewise, a high ROUGE score may reflect numerous overlapping n-grams but fail to capture true style transfer.

We introduce a new metric, "Register-Specific Words" (RSW), to evaluate a model's ability to convert words between informal and formal registers. RSW focuses on identifying and assessing the transformation of words with distinct formal and informal forms (e.g. خونه (pr: khune; tr: home) and its formal form خانه (pr: kha:ne)), excluding other words to measure true style transfer efficacy accurately. This involves identifying RSWs in the reference text, evaluating the generated text for accurate conversion of these RSWs, and normalizing this count against the total RSWs found in the reference to calculate a percentage-based score.

Additionally, we introduce a "Tag Matching Score" (TMS) to assess grammatical alignment between generated and reference formal sentences by comparing their part-of-speech tags. TMS calculates error rates based on mismatches, averages these across all sentences, and converts the result into a percentage of grammatical accuracy, with higher values indicating better alignment. We utilize Hazm's POS tagger for this purpose.

Finally, we compute the harmonic mean of the RSW score and TMS to calculate the "Formality Transfer Index' (FTI), providing a comprehensive measure that accounts for both style transfer accuracy and grammatical consistency. FTI surpasses traditional metrics like BLEU and ROUGE by directly assessing the nuances of linguistic style transformation and grammatical integrity, crucial for tasks such as Persian formality style transfer, where subtle stylistic nuances play a significant role.

## 6. Results

### 6.1 Model Performance Comparison

When benchmarked against baseline methods using different evaluation metrics, our model shows a distinct performance in both settings, word alignment dictionary and autonomous setting (Table 3). Our proposed Autonomous Fa-BERT2BERT model, incorporating consistency learning, outperforms all other methods with higher scores across all metrics: 70.68 in BLEU, 86.15 in ROUGE-L, and 75.83 in our newly proposed FTI score. This superior performance was also consistent when evaluated on short and long test sets separately (Table 4,5).

**Table 3** Comparative Results on Aggregated Test Set.

| Model | BLEU Score | Rouge-L | RSW Score | Tag Matching Score (TMS) | Formality Style Index(FTI) |
|---|---|---|---|---|---|
| Word-Alignment Dictionary | 22.87 | 52.22 | 61.82 | 32.24 | 42.38 |
| Fa-BERT2BERT With word alignment dictionary | 63.54 | 80.53 | 78.15 | 61.24 | 68.67 |
| **Autonomous Fa-BERT2BERT** | **70.68** | **86.15** | **81.10** | **71.12** | **75.83** |
| Pars-BERT2BERT With word alignment dictionary | 50.56 | 75.96 | 75.31 | 54.23 | 63.32 |
| Autonomous Pars-BERT2BERT | 65.83 | 84.89 | **82.52** | 69.43 | 75.41 |
| Hazm | 36.70 | 60.13 | 66.74 | 42.00 | 51.56 |
| MT5 Back-translation | 17.90 | 35.95 | 41.28 | 37.56 | 39.30 |
| Rule-based | 34.36 | 54.21 | 57.61 | 38.08 | 45.64 |

**Rule-based models**: Models such as those extending Armin and Shamsfard (2011) and the Hazm framework, face two main challenges. Firstly, they struggle with finding formal equivalents for words that defy rule-based categorization. For example, converting informal Persian words like "واسه" (pr: va:se; tr: for) or "خونه" (pr: khune; tr: home) to their formal counterparts "برای" (pr: bara:ye) and "خانه" (pr: kha:ne) cannot be governed by any specific rule. Secondly, they inadequately handle syntactic conversions, especially in reordering words to fit the formal SOV structure of contemporary Persian grammar (see examples in table 7). This is reflected in their low Tag Matching Scores (42.00 and 41.08) compared to relatively higher RSW scores (66.74 and 57.61)

When assessing these models on test sets with varying length, both rule-based models exhibited a distinct performance profile compared to other models. While other models often achieved higher scores on short set, rule-based models performed better scores on long set except for TMS. This could be attributed to their adeptness in handling repetitive structures in longer texts, leading to error dilution and better matching of significant text sequences, particularly for metrics like BLEU and ROUGE-L.

**Back-translation-based Model**: Employing back-translation through an intermediate language tends to fail to preserve the content of the formal reference, in both short and long length inputs. In most cases the problem is attributed to the model's inaccurate translation capabilities both into and from intermediate languages. In few cases, the generated output fails to match the expected formal wording, utilizing synonyms instead, which results in lower scores based on our benchmarks. Employing multiple reference texts for comparison could offer a more nuanced evaluation of model performance in such cases.

Table 4 Comparative Results on Short Test Set

| Model | BLEU Score | Rouge-L | RSW Score | Tag Matching Score (TMS) | Formality Style Index (FTI) |
|---|---|---|---|---|---|
| Autonomous Fa-BERT2BERT | **66.24** | **82.13** | 78.86 | **70.14** | **74.80** |
| Fa-BERT2BERT With word alignment dictionary | 53.24 | 76.37 | 72.24 | 54.95 | 61.77 |
| Autonomous Pars-BERT2BERT | 61.97 | 83.06 | **80.49** | 69.01 | 74.31 |
| Pars-BERT2BERT With word alignment dictionary | 51.18 | 76.97 | 75.77 | 56.74 | 64.89 |
| Hazm | 41.20 | 66.73 | 70.57 | 40.74 | 51.66 |

Table 5 Comparative Results on Long Test Set

| Model | BLEU Score | Rouge-L | RSW Score | Tag Matching Score (TMS) | Formality Style Index (FTI) |
|---|---|---|---|---|---|
| Autonomous Fa-BERT2BERT | **79.35** | 86.36 | 86.61 | **71.26** | **78.19** |
| Fa-BERT2BERT With word alignment dictionary | 59.17 | 76.81 | 73.23 | 56.67 | 63.89 |
| Autonomous Pars-BERT2BERT | 74.09 | **87.21** | **88.39** | 69.97 | 78.11 |
| Pars-BERT2BERT With word alignment dictionary | 57.22 | 78.03 | 79.60 | 55.71 | 65.55 |
| Hazm | 24.63 | 52.90 | 57.31 | 43.39 | 49.39 |

**Fa-BERT2BERT:** The first setting, where we used a dictionary-based method for lexical style conversion before training Fa-BERT2BERT, resulted in BLEU, ROUGE-L and FTI scores of 63.54, 80.53 and 68.67 respectively, a notable increase from the initial scores of 22.87 , 52.22 and 42.38 achieved with dictionary-based word replacement alone. Additionally, the RSW score improved from 61.82 to 78.15, and TMS from 32.24 to 61.24, indicating significant lexical and grammatical enhancements using Fa-BERT2BERT, especially in grammatical accuracy.

Conversely, the second setting, which harnessed the autonomous capabilities of the Fa-BERT2BERT architecture, showed even stronger results, with scores of 70.68, 86.15, 75.83, 71.12 for BLEU, Rouge-L and FTI respectively. In a comparative evaluation, the autonomous Fa-BERT2BERT setting yielded higher scores across all metrics, thereby validating its ability to effectively manage both lexical and syntactic style transformations without external aids. This suggests a superior efficacy of context-aware transformations over context-blind lexical substitutions.

When evaluating performance across different input lengths, it becomes evident that both models exhibit superior performance on shorter test sets. This improvement can largely be attributed to the limitations imposed by the context window length, which restricts the models' capability to effectively grasp and leverage extended contexts. Remarkably, the Fa-BERT2BERT model, enhanced with a word alignment dictionary, demonstrates more consistent performance across various sets, indicating a better robustness.

Through a more detailed examination of model performances, we can observe the models' limitations. The Fa-BERT2BERT model, when implementing the word alignment dictionary strategy, begins by replacing words without considering context. This approach can result in incorrect substitutions, where the replaced words, while individually accurate, may not align with the sentence's overall context. As a result, the original message provided to the transformer model may be distorted. For example the word "خونم" (pr: khunam) can mean both 'my house' and 'my blood' depending on the context and a wrong replacement can impact the meaning significantly.

In the autonomous setting of the Fa-BERT2BERT model, where the system self-learns to perform style conversion, there's an observed limitation in its handling of zero-width non-joiner (ZWNJ) characters, which are a significant feature in written Persian to distinguish words that should not be connected. The model occasionally replaces the ZWNJ with a regular whitespace, as seen in the transformation of " اطلاع رسونی" to "اطلاع رسانی" instead of the correct "اطلاع‌رسانی" (see example in table 6). This minor error can sometimes impact the meaning of the words, examples include "دانش آموز" (pr: dānesh āmooz; tr: learn knowledge) and "دانش‌آموز" (pr: dānesh āmooz; tr: student), and "یکجا" (pr: yekjā, tr: altogether) versus "یک جا" (pr: yek jā, tr: a place)

Finally, all models showed limitations in grasping stylistic nuances, particularly idioms and proverbs deeply embedded in cultural contexts. These expressions carry meanings beyond their literal words, but the models often only provide literal translations, missing metaphorical and cultural nuances. This issue arises because these nuances are not only linguistic but cultural, and if the training data lacks these expressions, models struggle with accurate interpretation.

**ParsBERT2BERT:** The Pars-BERT2BERT model exhibits notable improvements when augmented with a word alignment dictionary, achieving a BLEU score of 50.56 and a ROUGE-L score of 75.96, suggesting enhanced translation quality over basic methods. However, the autonomous "Pars-BERT2BERT" model surpasses this with a BLEU score of 65.83 and a ROUGE-L score of 84.89, highlighting its superior capability in handling complex linguistic transformations without external aids.

Overall ParsBERT2BERT models show a performance profile similar to Fa-BERT2BERT models across varying length of input data, but with the latter demonstrating better performance, as reflected in the comparative scores. This highlights the effectiveness of Fa-BERT compared to ParsBERT in style transfer. Like Fa-BERT2BERT, the Pars-BERT2BERT models also struggle with the correct handling of zero-width non-joiner (ZWNJ) characters and conveying the meaning of idiomatic expressions and proverbs.

**Table 6** a comparison of models for an informal text. In the informal sentence, informal words are in red and in the formal reference and generated sentences, accurately converted words are marked in green, while errors are indicated in red. The orange colored words show minor errors; those converted correctly to formal words but their writing style is incorrect as the two parts of a compound word are separated by space instead of a ZWNJ character.

| Model | Sentence |
|---|---|
| Informal | دمشون گرم اطلاع رسونی کردن |
| Formal reference | دمشان گرم که اطلاع‌رسانی کردند |
| Autonomous Fa-BERT2BERT | دمشان گرم که اطلاع رسانی کردند |
| Fa-BERT2BERT with word alignment dictionary | دمشان گرم اطلاع‌رسانی  کردند. |
| Autonomous Pars-BERT2BERT | دمشان گرم که اطلاع رسانی کردند |
| Pars-BERT2BERT with word alignment dictionary | دمشان گرم اطلاع‌رسانی  کردند |
| Hazm | دمشان گرم اطلاع رسانی کردند |
| Rule-based | دمشون گرم اطلاع رسونی کردند |

In this example, both rule-based models were unable to accurately convert the informal term "اطلاع رسونی" (pr: etelāʼ resuni; tr: informing) to its formal equivalent "اطلاع‌رسانی" (pr: etelāʼresāni; tr: information dissemination). Additionally, both Autonomous models, while generally effective in other areas, failed in correctly generating the zero-width non-joiner (ZWNJ) between "اطلاع" (pr: etelāʼ) and "رسانی" (pr: resāni).

**Table 7** An example of an informal text and comparison of models performance. In the informal sentence, informal words are shown in red and the part of the sentence which needs movements and syntactic reordering (informal

grammatical structure) is highlighted in red. In the formal reference and generated sentences, accurately converted words are written in green, while errors are indicated in red. Additionally, green highlights show the correct word reorderings and red highlights show incorrect reorderings.

| Model | Sentence |
|---|---|
| Informal | سیاهی گناه پاک میشه از وجودمون و نور تو قلب مون بیشتر و بیشتر میشه. |
| Formal reference | سیاهی گناه از وجودمان پاک می شود و نور در قلبمان بیشتر و بیشتر می‌شود |
| Autonomous Fa-BERT2BERT | سیاهی گناه از وجودمان پاک می شود و نور در قلبمان بیشتر و بیشتر می‌شود. |
| Fa-BERT2BERT With word alignment dictionary | سیاهی گناه از وجودمان پاک می شود و نور تو قلب ما بیشتر و بیشتر می‌شود. |
| Autonomous Pars-BERT2BERT | سیاهی گناه از وجود ما پاک می شود و نور در قلبمان بیشتر و بیشتر می‌شود. |
| Pars-BERT2BERT With word alignment dictionary | سیاهی گناه از وجودم پاک می شود و نور تو قلب مان بیشتر و بیشتر می‌شود. |
| Hazm | سیاهی گناه پاک می شود از وجودمان و نور تو قلبمان بیشتر و بیشتر می‌شود. |
| Rule-based | سیاهی گناه پاک می شود از وجودمان و نور در قلبمان بیشتر و بیشتر می‌شود. |

As shown in table 7, the autonomous models successfully converted the input to its formal form, accurately addressing both grammatical and linguistic changes. In contrast, the both ParsBert2Bert and Fa-BERT2BERT models with a word alignment dictionary correctly applied grammatical adjustments but inaccurately translated "تو" (pr:tū; tr:"in") to "in", likely due to the context-insensitive nature of the word alignment approach. Additionally, the ParsBERT2BERT with word alignment has mistakenly converted the word "وجودمان" (pr: vodʒuːdemaːn; tr: our existence) to "وجودم"(pr: vodʒuːdam; tr: my existence). The rule-based models, while adept at implementing linguistic modifications, fell short in both converting "تو" to "در" as this is an irregular word and there is no rule to govern these conversions and also rearranging word order to align with formal grammatical structures.

## 6.2 Comparing Training Strategies

The results presented in table 8 compare the proposed Autonomous Fa-BERT2BERT model trained with this gradient-based consistency learning approach to the same model trained using a standard approach. The model utilizing the new strategy outperforms the standard in every measured aspect. Notably, the Tag Matching Score (TMS), which assesses grammatical adherence, sees the largest improvement, indicating that the training strategy effectively enhances grammatical consistency in the style transfer process. Additionally, other metrics such as BLEU score, Rouge-L, RSW Score, and Formality Style Index (FTI) also exhibit notable improvements, underscoring the comprehensive enhancement in the model's performance due to the new training strategy.

**Table 8** Comparative Performance of Fa-BERT2BERT Models: Impact of Gradient-Based Consistency Learning on Formality Style Transfer Metrics

| Model | BLEU Score | Rouge-L | RSW Score | Tag Matching Score (TMS) | Formality Style Index (FTI) |
|---|---|---|---|---|---|
| Autonomous Fa-BERT2BERT with Gradient-based Consistency Learning Approach | **70.68** | **86.15** | **81.10** | **71.12** | **75.83** |
| Fa-BERT2BERT with Standard training Approach | 63.54 | 80.53 | 78.15 | 61.24 | 68.67 |

# 7. Conclusion and Future Work

In this paper we propose the novel Fa-BERT2BERT model, which demonstrates superior performance in formality style transfer, particularly in handling the complexities of Persian text. Our innovative use of consistency learning and gradient-based dynamic weighting for training has proven effective in navigating the syntactic and lexical intricacies, thereby enhancing the accuracy and functionality of NLP applications in transforming informal Persian text to its formal counterpart.

Future research will focus on overcoming the limitations of the current model, particularly in its handling of the zero-width non-joiner (ZWNJ) space, which is crucial for accurate Persian text processing. Additionally, improving the model's capability to interpret and convert slangs, idioms, and cultural expressions prevalent in informal Persian will be a priority, as these elements are integral to capturing the language's nuance. Exploring the potential of multilingual Large Language Models (LLMs) will also be beneficial, aiming to analyze their effectiveness and adaptability to the specific challenges of Persian formality style transfer.

# 8. Statements and Declarations

## 8.1 Data Availability Statement

This study utilized ParsMap, an Informal-Formal Persian Corpus, a parallel freely available corpus for informal and formal Persian language, as introduced by Tajalli et al. [2]. The models are also publicly available at the author's [Hugging Face](#) account.

## 8.2 Competing Interests

All authors certify that they have no affiliations with or involvement in any organization or entity with any financial interest or non-financial interest in the subject matter or materials discussed in this manuscript.